\documentclass[10pt]{article}

\usepackage{times}
\usepackage{epsfig}
\usepackage{graphicx}
\usepackage{amsmath}
\usepackage{amssymb}
\usepackage{tiberio}

\usepackage[pagebackref=true,breaklinks=true,letterpaper=true,colorlinks,bookmarks=false]{hyperref}

\begin{document}

\title{Exponential Family Graph Matching and Ranking}
\author{James Petterson, Tib\'erio Caetano, Julian McAuley, Jin Yu
\thanks{NICTA's Statistical Machine Learning program, Locked Bag 8001, ACT 2601, Australia, and Research School of Information Sciences and Engineering, Australian National University, ACT 0200, Australia. NICTA is funded by the Australian Government'Õs Backing Australia'Õs Ability initiative, and the Australian Research CouncilÕ's ICT Centre of Excellence program. e-mails: \texttt{first.last@nicta.com.au}}}

\maketitle

\begin{abstract}
We present a method for learning max-weight matching predictors in bipartite graphs. The method consists of performing maximum a posteriori estimation in exponential families with sufficient statistics that encode permutations and data features. Although inference is in general hard, we show that for one very relevant application--web page ranking--exact inference is efficient. For general model instances, an appropriate sampler is readily available. Contrary to existing max-margin matching models, our approach is statistically consistent and, in addition, experiments with increasing sample sizes indicate superior improvement over such models. We apply the method to graph matching in computer vision as well as to a standard benchmark dataset for learning web page ranking, in which we obtain state-of-the-art results, in particular improving on max-margin variants. The drawback of this method with respect to max-margin alternatives is its runtime for large graphs, which is comparatively high.
\end{abstract}

\section{Introduction}
The Maximum-Weight Bipartite Matching Problem (henceforth `matching problem') is a fundamental problem in combinatorial optimization \cite{PapSte82}. This is the problem of finding the `heaviest' perfect match in a weighted bipartite graph. An exact optimal solution can be found in cubic time by standard methods such as the Hungarian algorithm.

This problem is of practical interest because it can nicely model real-world applications. For example, in computer vision the crucial problem of finding a correspondence between sets of image features is often modeled as a matching problem \cite{Belongie02,CaeMcACheLeSmo09}. Ranking algorithms can be based on a matching framework \cite{LeSmo07}, as can clustering algorithms \cite{JebSch06,HuaJeb07}.

When modeling a problem as one of matching, one central question is the choice of the weight matrix. The problem is that in real applications we typically observe edge \emph{feature vectors}, not edge weights. Consider a concrete example in computer vision: it is difficult to tell what the `similarity score' is between two image feature points, but it is straightforward to extract feature vectors (e.g.~SIFT) associated with those points. 

In this setting, it is natural to ask whether we could parameterize the features, and use \emph{labeled matches} in order to estimate the parameters such that, given graphs with `similar' features, their resulting max-weight matches are also `similar'. This idea of `parameterizing algorithms' and then optimizing for agreement with data is called \emph{structured estimation} \cite{Taskar04,TsoJoaHofAlt05}. 

\cite{Taskar04} and \cite{CaeMcACheLeSmo09} describe max-margin structured estimation formalisms for this problem. Max-margin structured estimators are appealing in that they \emph{try} to minimize the loss that one really cares about (`structured losses', of which the Hamming loss is an example). However structured losses are typically piecewise constant in the parameters, which eliminates any hope of using smooth optimization directly. Max-margin estimators instead minimize a surrogate loss which is easier to optimize, namely a convex upper bound on the structured loss \cite{TsoJoaHofAlt05}. In practice the results are often good, but known convex relaxations produce estimators which are statistically inconsistent \cite{McAllester07}, i.e.,~the algorithm in general fails to obtain the best attainable model in the limit of infinite training data. The inconsistency of multiclass support vector machines is a well-known issue in the literature that has received careful examination recently \cite{LiuSheDos05,LiuShe06}.

Motivated by the inconsistency issues of max-margin structured estimators as well as by the well-known benefits of having a full probabilistic model, in this paper we present a maximum a posteriori (MAP) estimator for the matching problem. The observed data are the edge feature vectors and the labeled matches provided for training. We then maximize the conditional posterior likelihood of matches given the observed data. We build an exponential family model where the sufficient statistics are such that the mode of the distribution (the prediction) is the solution of a max-weight matching problem. The resulting partition function is $\sharp$P-complete to compute exactly. However, we show that for \emph{learning to rank} applications the model instance is tractable. We then compare the performance of our model instance against a large number of state-of-the-art ranking methods, including DORM \cite{LeSmo07}, an approach that only differs to our model instance by using max-margin instead of a MAP formulation. We show very competitive results on standard webpage ranking datasets, and in particular we show that our model performs better than or on par with DORM. For intractable model instances, we show that the problem can be approximately solved using sampling and we provide experiments from the computer vision domain. However the fastest suitable sampler is still quite slow for large models, in which case max-margin matching estimators like those of \cite{CaeMcACheLeSmo09} and \cite{Taskar04} are likely to be preferable even in spite of their potential inferior accuracy.

\section{Background}

\subsection{Structured Prediction}

In recent years, great attention has been devoted in Machine Learning to so-called \emph{structured predictors}, which are predictors of the kind
\begin{align}
\label{eq:predictor}
g_{\theta}:\Xcal \mapsto \Ycal,
\end{align}
where $\Xcal$ is an arbitrary input space and $\Ycal$ is an \emph{arbitrary discrete space, typically exponentially large}. $\Ycal$ may be, for example, a space of matrices, trees, graphs, sequences, strings, matches, etc. This structured nature of $\Ycal$ is what \emph{structured prediction} refers to. In the setting of this paper, $\Xcal$ is the set of vector-weighted bipartite graphs (i.e.,~each edge has a feature vector associated to it), and $\Ycal$ is the set of perfect matches induced by $\Xcal$. If $N$ graphs are available, along with corresponding annotated matches (i.e.,~a set $\{(x^n,y^n)\}^N_{n=1}$), our task will be to \emph{estimate} $\theta$ such that when we apply the predictor $g_{\theta}$ to a new graph it produces a match that is similar to matches of similar graphs from the annotated set. \emph{Structured learning} or \emph{structured estimation} refers to the process of estimating a vector $\theta$ for predictor $g_{\theta}$ when data $\{(x^1,y^1),\dots,(x^N,y^N)\}\in (\Xcal \times \Ycal)^N$ are available. \emph{Structured prediction} for input $x$ means computing $y=g(x;\theta)$ using the estimated $\theta$.

Two generic estimation strategies have been popular in producing structured predictors. One is based on max-margin estimators \cite{TsoJoaHofAlt05,TasGueKol04,Taskar04}, and the other on maximum-likelihood (ML) or MAP estimators in exponential family models \cite{LafMcCPer01}.

The first approach is a generalization of support vector machines to the case where the set $\Ycal$ is structured. However the resulting estimators are known to be \emph{inconsistent} in general: in the limit of infinite training data the algorithm fails to recover the best model in the model class \cite{McAllester07,LiuShe06,LiuSheDos05}. McAllester recently provided an interesting analysis on this issue, where he proposed new upper bounds whose minimization results in consistent estimators, but no such bounds are convex \cite{McAllester07}. 
The other approach uses ML or MAP estimation in conditional exponential families with `structured' sufficient statistics, such as in probabilistic graphical models, where they are decomposed over the cliques of the graph (in which case they are called Conditional Random Fields, or CRFs \cite{LafMcCPer01}). In the case of tractable graphical models, dynamic programming can be used to efficiently perform inference. Other tractable models of this type include models that predict spanning trees and models that predict binary labelings in planar graphs \cite{GloJaa07,KooGloCarCol07}. ML and MAP estimators in exponential families not only amount to solving an unconstrained and convex optimization problem; in addition they are statistically consistent. The main problem with these types of models is that often the partition function is intractable. This has motivated the use of max-margin methods in many scenarios where such intractability arises. 

\subsection{The Matching Problem}

Consider a weighted bipartite graph with $m$ nodes in each part, $G=(V,E,w)$, where $V$ is the set of vertices, $E$ is the set of edges and $w:E \mapsto \mathbb R$ is a set of real-valued weights associated to the edges. $G$ can be simply represented by a matrix ($w_{ij}$) where the entry $w_{ij}$ is the weight of the edge $ij$. Consider also a bijection $y:\{1,2,\dots,m\}\mapsto \{1,2,\dots,m\}$, i.e.,~a permutation. Then the matching problem consists of computing

\begin{align}
\label{eq:lap}
y^* = \argmax_{y} \sum_{i=1}^m w_{iy(i)}.
\end{align}
This is a well-studied problem; it is tractable and can be solved in $O(m^3)$ time \cite{JonVol87,PapSte82}. This model can be used to match features in images \cite{CaeMcACheLeSmo09}, improve classification algorithms \cite{HuaJeb07} and rank webpages \cite{LeSmo07}, to cite a few applications. The typical setting consists of engineering the score matrix $w_{ij}$ according to domain knowledge and subsequently solving the combinatorial problem.

\section{The Model}

\subsection{Basic Goal}
\begin{figure}[tb]
\centerline{\includegraphics[width=0.5\textwidth]{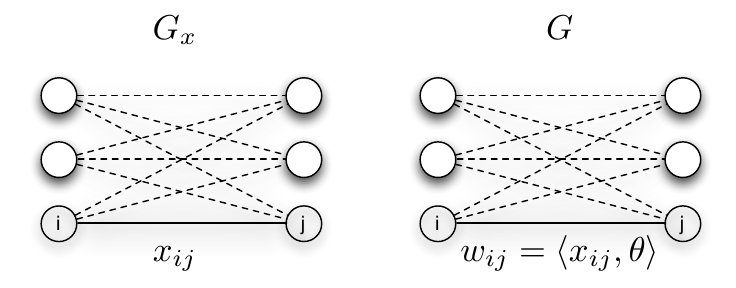}}
\caption{ Left: Illustration of an input \emph{vector-weighted} bipartite graph $G_{x}$ with $3\times3$ edges. There is a vector $x_{e}$ associated to each edge $e$ (for clarity only $x_{ij}$ is shown, corresponding to the solid edge). Right: weighted bipartite graph $G$ obtained by evaluating $G_{x}$ on the learned vector $\theta$ (again only edge $ij$ is shown). }
\label{fig:graphs}
\end{figure}

In this paper we assume that the weights $w_{ij}$ are instead to be \emph{estimated} from training data. More precisely, the weight $w_{ij}$ associated to the edge $ij$ in a graph will be the result of an appropriate composition of a \emph{feature vector} $x_{ij}$ (observed) and a \emph{parameter vector} $\theta$ (estimated from training data). Therefore, in practice, our input is a \emph{vector-weighted} bipartite graph $G_{x}=(V,E,x)$ ($x:E\mapsto \mathbb R^n$), which is `evaluated' at a particular $\theta$ (obtained from previous training) so as to attain the graph $G=(V,E,w)$. See Figure \ref{fig:graphs} for an illustration.

More formally, assume that a training set $\{X,Y\}=\{(x^n,y^n)\}^N_{n=1}$ is available, where $x^n:=(x^n_{11},x^n_{12}\dots,x^n_{M(n)M(n)})$. Here $M(n)$ is the number of nodes in each part of the vector-weighted bipartite graph $x^n$. We then parameterize $x_{ij}$ as $w_{iy(i)}=f(x_{iy(i)};\theta)$, and the goal is to find the $\theta$ which maximizes the posterior likelihood of the observed data. We will assume $f$ to be bilinear, i.e.,~$f(x_{iy(i)};\theta)=\inner{x_{iy(i)}}{\theta}$.

\subsection{Exponential Family Model}
We assume an exponential family model, where the probability model is
\begin{align}
p(y|x;\theta)=\exp{\left(\inner{\phi(x,y)}{\theta}-g(x;\theta)\right)}, \text{ where}
\end{align}
\begin{align}
g(x;\theta)=\log \sum_{y} \exp{\inner{\phi(x,y)}{\theta}}
\end{align}
is the log-partition function, which is a convex and differentiable function of $\theta$ \cite{WaiJor03}. 

The prediction in this model is the most likely $y$, i.e.,
\begin{align}
\label{eq:argmax}
y^*=\argmax_{y} p(y|x;\theta)=\argmax_{y} \inner{\phi(x,y)}{\theta}
\end{align}
and ML estimation amounts to maximizing the conditional likelihood of the training set $\{X,Y\}$, i.e.,~computing $\argmax_{\theta} p(Y|X;\theta)$. In practice we will in general introduce a prior on $\theta$ and perform MAP estimation:
\begin{align}
\theta^* = \argmax_{\theta} p(Y|X;\theta)p(\theta)=\argmax_{\theta}p(\theta|Y,X).
\end{align}
Assuming iid sampling, we have $p(Y|X;\theta)=\prod_{n=1}^N p(y^n|x^n;\theta)$. Therefore,
\begin{align}
& p(\theta|Y,X)  \propto p(\theta)\prod_{n=1}^N \exp{\left(\inner{\phi(x^n,y^n)}{\theta}-g(x^n;\theta)\right)}
\nonumber
\\
= & \exp{\left(\log p(\theta)+\sum_{n=1}^N \left(\inner{\phi(x^n,y^n)}{\theta}-g(x^n;\theta)\right)\right)}.
\end{align}

We impose a Gaussian prior on $\theta$. Instead of maximizing the posterior we can instead minimize the negative log-posterior $\ell(Y|X;\theta)$, which becomes our loss function (we suppress the constant term):
\begin{align}
\label{eq:loss}
\ell(Y|X;\theta) = & \frac{\lambda}{2}\nbr{\theta}^2 + \frac{1}{N}\sum_{n=1}^N \left(g(x^n;\theta) - \inner{\phi(x^n,y^n)}{\theta}\right)
\end{align}
where $\lambda$ is a regularization constant. $\ell(Y|X;\theta)$ is a convex function of $\theta$ since the log-partition function $g(\theta)$ is a convex function of $\theta$ \cite{WaiJor03} and the other terms are clearly convex in $\theta$.

\subsection{Feature Parameterization}
The critical observation now is that we equate the solution of the matching problem \eq{eq:lap} to the prediction of the exponential family model \eq{eq:argmax}, i.e.,~$\sum_{i}w_{iy(i)}=\inner{\phi(x,y)}{\theta}$. Since our goal is to parameterize features of individual pairs of nodes (so as to produce the weight of an edge), the most natural model is
\begin{align}
\phi(x,y) & =  \sum_{i=1}^M x_{iy(i)}, \text{ which gives }\\
\label{eq:param}
w_{iy(i)} & =  \inner{x_{iy(i)}}{\theta},
\end{align}
i.e.,~linear in both $x$ and $\theta$ (see Figure \ref{fig:graphs}, right). The specific form for $x_{ij}$ will be discussed in the experimental section. In light of \eq{eq:param}, \eq{eq:lap} now clearly means a \emph{prediction} of the best match for $G_{x}$ under the model $\theta$.
\section{Learning the Model}
\label{sec:learning}
\subsection{Basics}
\label{sec:perm}
We need to solve $\theta^* = \argmin_{\theta} \ell(Y|X;\theta)$. $\ell(Y|X;\theta)$ is a convex and differentiable function of $\theta$ \cite{WaiJor03}, therefore gradient descent will find the global optimum. In order to compute $\nabla_{\theta}\ell(Y|X;\theta)$, we need to compute $\nabla_{\theta}g(\theta)$. It is a standard result of exponential families that the gradient of the log-partition function is the expectation of the sufficient statistics:
\begin{align}
\label{eq:grad}
\nabla_{\theta}g(x;\theta) = \Eb_{y\sim p(y|x;\theta)}[\phi(x,y)].
\end{align}
Therefore in order to perform gradient descent we need to compute the above expectation. Opening the above expression gives 
\begin{align}
\label{eq:expectation1}
& \Eb_{y\sim p(y|x;\theta)}[\phi(x,y)]=\sum_{y}\phi(x,y) p(y|x;\theta)\\
\label{eq:expectation2}
& = \frac{1}{Z(x;\theta)}\sum_{y}\phi(x,y) \prod_{i=1}^M \exp(\inner{x_{iy(i)}}{\theta}),
\end{align}
which reveals that the partition function $Z(x;\theta)$ needs to be computed. The partition function is:
\begin{align}
\label{eq:partition}
Z(x;\theta) = \sum_{y} \prod_{i=1}^M \underbrace{\exp(\inner{x_{iy(i)}}{\theta})}_{=:B_{iy(i)}}.
\end{align}
Note that the above is the expression for the \emph{permanent} of matrix $B$ \cite{Minc78}. The permanent is similar in definition to the determinant, the difference being that for the latter $\sgn(y)$ comes before the product. However, unlike the determinant, which is computable efficiently and exactly by standard linear algebra manipulations \cite{KooGloCarCol07}, computing the permanent is a $\sharp$P-complete problem \cite{Valiant79}. Therefore we have no realistic hope of computing \eq{eq:grad} exactly for general problems. 

\subsection{Exact Expectation}
\label{sec:exact_exp}

The exact partition function itself can be efficiently computed for up to about $M=30$ using the $O(M2^M)$ algorithm by Ryser \cite{Ryser63}. However for arbitrary expectations we are not aware of any exact algorithm which is more efficient than full enumeration (which would constrain tractability to very small graphs). However we will see that even in the case of very small graphs we find a very important application: learning to rank. In our experiments, we successfully apply a tractable instance of our model to benchmark webpage ranking datasets, obtaining very competitive results. For larger graphs, we have alternative options as indicated below.

\subsection{Approximate Expectation}
\label{sec:approx_exp}
If we have a situation in which the set of feasible permutations is too large to be fully enumerated efficiently, we need to resort to some approximation for the expectation of the sufficient statistics. The best solution we are aware of is one by Huber and Law, who recently presented an algorithm to approximate the permanent of dense non-negative matrices \cite{HubLaw08}. The algorithm works by producing \emph{exact samples} from the distribution of perfect matches on weighted bipartite graphs. This is in precisely the same form as the distribution we have here, $p(y|x;\theta)$ \cite{HubLaw08}. We can use this algorithm for applications that involve larger graphs.\footnote{The algorithm is described in the appendix.} We generate $K$ samples from the distribution $p(y|x;\theta)$, and directly approximate \eq{eq:expectation1} with a Monte Carlo estimate
\begin{align}
\Eb_{y\sim p(y|x;\theta)}[\phi(x,y)]\approx \frac{1}{K}\sum_{i=1}^K \phi(x,y_i).
\end{align}
In our experiments, we apply this algorithm to an image matching application.

\section{Experiments}

\subsection{Ranking}
Here we apply the general matching model introduced in previous sections to the task of \emph{learning to rank}. Ranking is a fundamental problem with applications in diverse areas such as document retrieval, recommender systems, product rating and others. We focus on web page ranking.

We are given a set of queries $\{q_k\}$ and, for each query $q_k$, a list of $D(k)$ documents $\{d^k_1,\dots,d^k_{D(k)}\}$ with corresponding ratings $\{r^k_1,\dots,r^k_{D(k)}\}$ (assigned by a human editor), measuring the relevance degree of each document with respect to query $q_k$. A rating or relevance degree is usually a nominal value in the list $\{1,\dots,R\}$, where $R$ is typically between 2 and 5. We are also given, for every retrieved document $d^k_i$, a joint feature vector $\psi^k_i$ for that document and the query $q_k$.

\textbf{Training} At training time, we model each query $q_k$ as a vector-weighted bipartite graph (Figure \ref{fig:graphs}) where the nodes on one side correspond to \emph{a subset} of cardinality $M$ of all $D(k)$ documents retrieved by the query, and the nodes on the other side correspond to all possible ranking positions for these documents (${1,\dots,M}$). The subset itself is chosen randomly, provided at least one exemplar document of every rating is present. Therefore $M$ must be such that $M\ge R$.

The process is then repeated in a bootstrap manner: we resample (with replacement) from the set of documents $\{d^k_1,\dots,d^k_{D(k)}\}$, $M$ documents at a time (conditioned on the fact that at least one exemplar of every rating is present, but otherwise randomly). This effectively boosts the number of training examples since each query $q_k$ ends up being selected many times, each time with a different subset of $M$ documents from the original set of $D(k)$ documents. 

In the following we drop the query index $k$ to examine a single query. Here we follow the construction used in \cite{LeSmo07} to map matching problems to ranking problems (indeed the only difference between our ranking model and that of \cite{LeSmo07} is that they use a max-margin estimator and we use MAP in an exponential family.) Our edge feature vector $x_{ij}$ will be the product of the feature vector $\psi_i$ associated with document $i$, and a scalar $c_j$ (the choice of which will be explained below) associated with ranking position $j$
\begin{align}
\label{eq:psi}
x_{ij} = \psi_i c_j.
\end{align}
$\psi_i$ is dataset specific (see details below).
From \eq{eq:param} and \eq{eq:psi}, we have $w_{ij} = c_j \inner{\psi_i}{\theta}$, 
and training proceeds as explained in Section \ref{sec:learning}.

\textbf{Testing} At test time, we are given a query $q$ and its corresponding list of $D$ associated documents. We then have to solve the prediction problem, i.e., 
\begin{align}
\label{eq:argmax_ranking1}
 y^* = & \argmax_{y} \sum_{i=1}^D \inner{x_{iy(i)}}{\theta} =
\argmax_{y} \sum_{i=1}^D c_{y(i)} \inner{\psi_i}{\theta}.
\end{align}
We now notice that if the scalar $c_j = c(j)$, where $c$ is a non-increasing function of rank position $j$, then \eq{eq:argmax_ranking1} can be solved simply by sorting the values of $\inner{\psi_i}{\theta}
$ in decreasing order.\footnote{If $r(v)$ denotes the vector of ranks of entries of vector $v$, then $\inner{a}{\pi(b)}$ is maximized by the permutation $\pi^*$ such that $r(a)=r(\pi^*(b))$, a theorem due to Polya, Littlewood, Hardy and Blackwell \cite{Sherman51}.} In other words, the matching problem becomes one of \emph{ranking the values $\inner{\psi_i}{\theta}$}. Inference in our model is therefore very fast (linear time).\footnote{Sorting the top $k$ items of a list of $D$ items takes $O(k\log k + D)$ time \cite{Martinez04}.} 
In this setting it makes sense to interpret the quantity $\inner{\psi_i}{\theta}$ as a \emph{score} of document $d_i$ for query $q$. This leaves open the question of which non-increasing function $c$ should be used. We do not solve this problem in this paper, and instead choose a fixed $c$. In theory it is possible to optimize over $c$ during learning, but in that case the optimization problem would no longer be convex. We describe the results of our method on LETOR 2.0 \cite{LiuXuQinXioetal07}, a publicly available benchmark data collection for comparing learning to rank algorithms. It is comprised of three data sets: OHSUMED, TD2003 and TD2004.

\textbf{Data sets} OHSUMED contains features extracted from query-document pairs in the OHSUMED collection, a subset of MEDLINE, a database of medical publications. It contains 106 queries. For each query there are a number of associated documents, with relevance degrees judged by humans on three levels: \emph{definitely}, \emph{possibly} or \emph{not relevant}. Each query-document pair is associated with a 25 dimensional feature vector, $\psi_i$. The total number of query-document pairs is 16,140. TD2003 and TD2004 contain features extracted from the topic distillation tasks of TREC 2003 and TREC 2004, with 50 and 75 queries, respectively. Again, for each query there are a number of associated documents, with relevance degrees judged by humans, but in this case only two levels are provided: \emph{relevant} or \emph{not relevant}. Each query-document pair is associated with a 44 dimensional feature vector,  $\psi_i$. The total number of query-document pairs is 49,171 for TD2003 and 74,170 for TD2004. All datasets are already partitioned for 5-fold cross-validation. See \cite{LiuXuQinXioetal07} for more details.

\textbf{Evaluation Metrics} In order to measure the effectiveness of our method we use the \emph{normalized discount cumulative gain} (NDCG) measure \cite{JarKek02b} at rank position $k$, which is defined as 
\begin{align}
\label{eq:ndcg}
\text{\emph{NDCG@k}} = \frac{1}{Z} \sum_{j=1}^k \frac{2^{r(j)}-1}{\log(1+j)},
\end{align}
where $r(j)$ is the relevance of the $j^{th}$ document in the list, and $Z$ is a normalization constant so that a perfect ranking yields an NDCG score of $1$.
\vspace{3mm}
\begin{figure*}[t]
\centerline{
\includegraphics[width=0.4\textwidth]{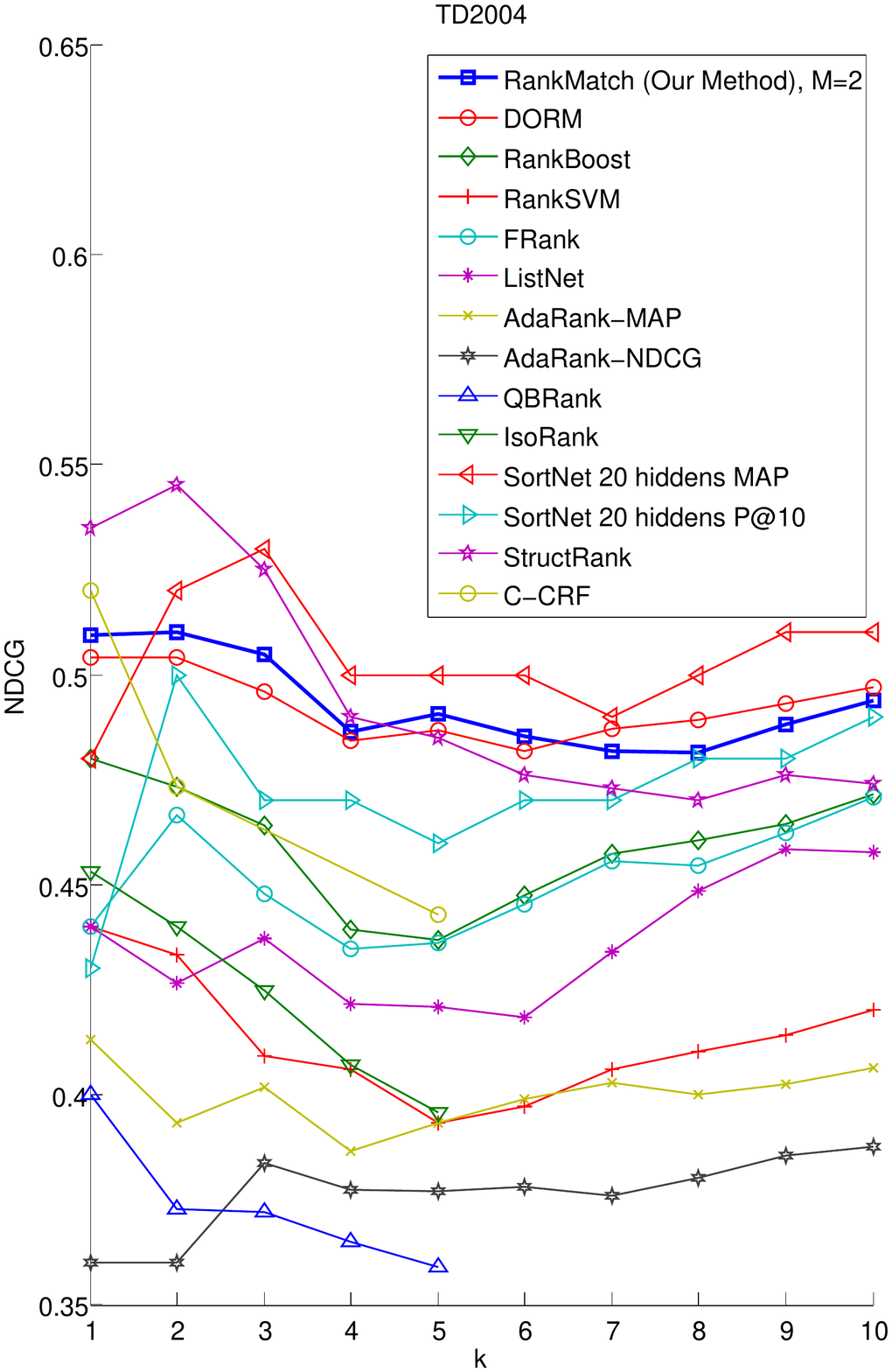}  
\hspace{-13mm}
\includegraphics[width=0.4\textwidth]{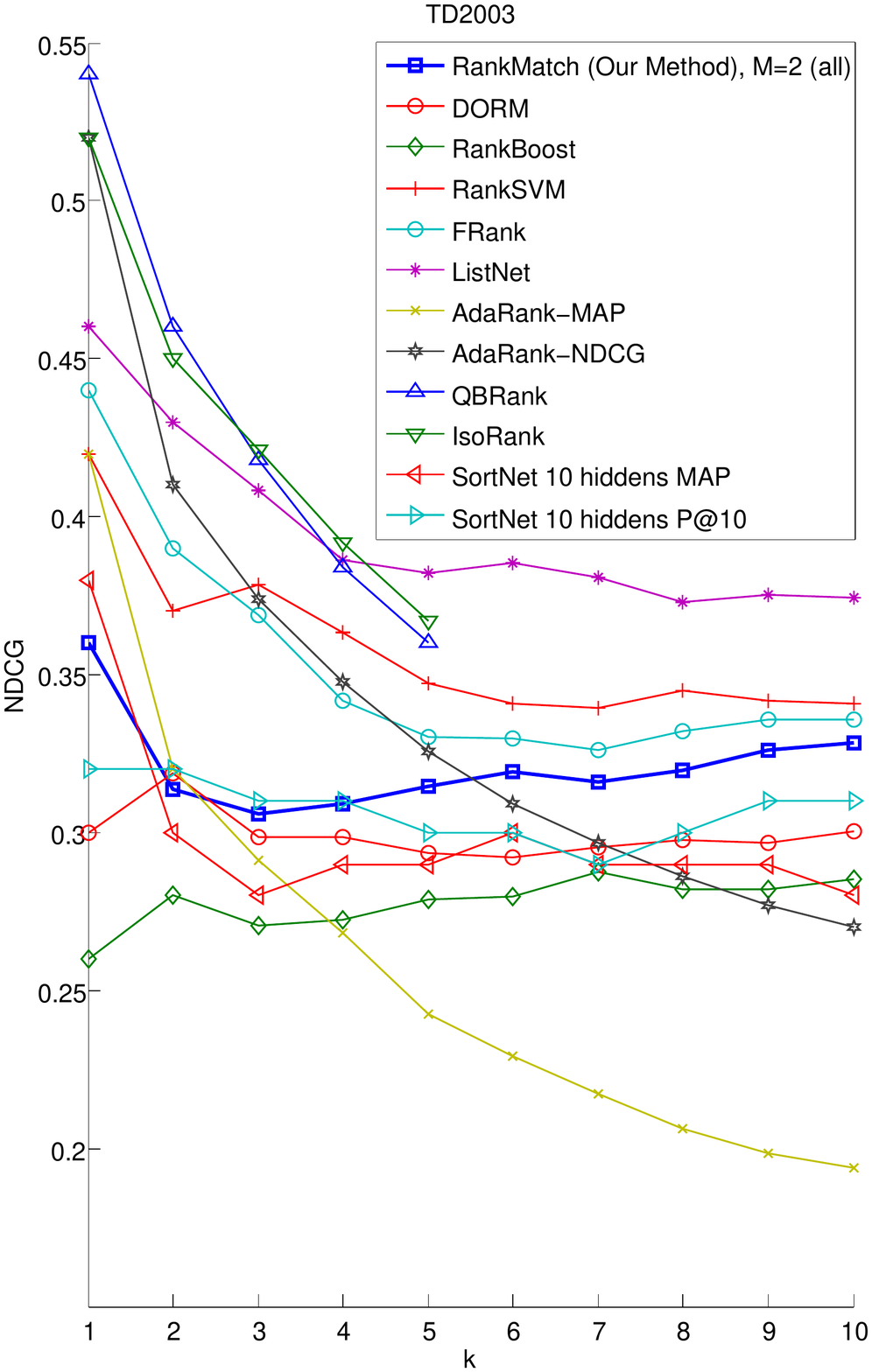}  
\hspace{-13mm}
\includegraphics[width=0.4\textwidth]{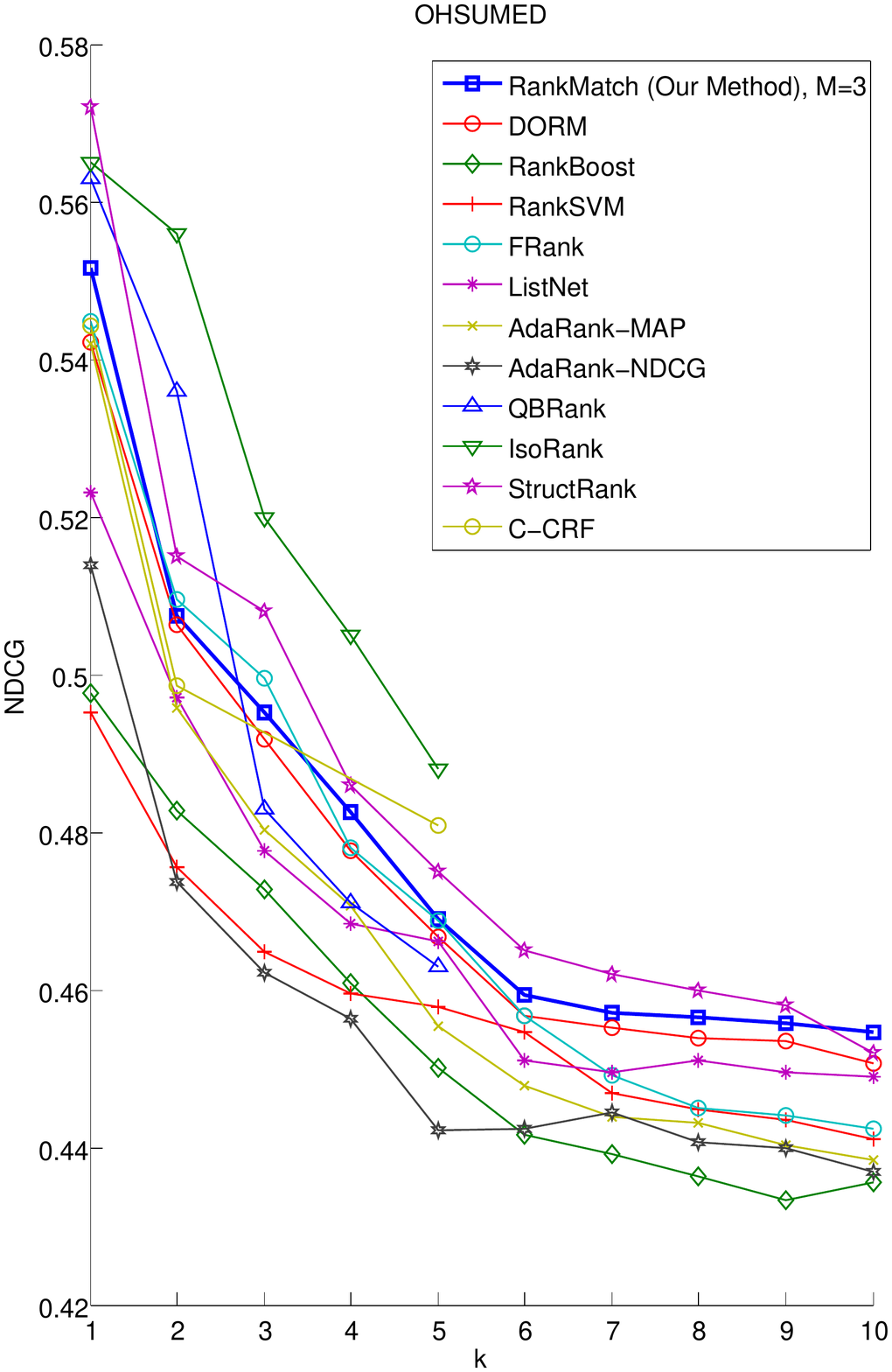}  
}
\vspace{-7mm}
\caption{Results of NDCG@k for state-of-the-art methods on TD2004 (left), TD2003 (middle) and OHSUMED (right). This is best viewed in color.}
\label{fig:comparative}
\end{figure*}

\textbf{External Parameters}
The regularization constant $\lambda$ is chosen by 5-fold cross-validation, with the partition provided by the LETOR package. All experiments are repeated 5 times to account for the randomness of the sampling of the training data. We use $c(j)=M-j$ on all experiments.

\textbf{Optimization} To optimize \eq{eq:loss} we use a standard BFGS Quasi-Newton method with a backtracking line search, as described in \cite{NocWri99}.

\textbf{Results} For the first experiment training was done on subsets sampled as described above, where for each query $q_k$  we sampled $0.4 \cdot D(k) \cdot M$ subsets, therefore increasing the number of samples linearly with $M$. For TD2003 we also trained with all possible subsets ($M=2 (all)$ in the plots). In Figure \ref{fig:comparative} we plot the results of our method (named RankMatch), for $M=R$, compared to those achieved by a number of state-of-the-art methods which have published NDCG scores in at least two of the datasets: RankBoost \cite{FreIyeSchSin03}, RankSVM \cite{HerGraObe00}, FRank \cite{TsaLiuQinCheetal07}, ListNet \cite{CaoQinLiyTsaetal07}, AdaRank \cite{XuLi07}, QBRank \cite{ZheZhaZhaChaetal08}, IsoRank \cite{ZheZhaSun08}, SortNet \cite{RigPapMagSca08}, StructRank \cite{HuaFre08} and C-CRF \cite{QinLiuZhaWanetal08}. We also included a plot of our implementation of DORM \cite{LeSmo07}, using \emph{precisely} the same resampling methodology and data for a fair comparison. RankMatch performs among the best methods on both TD2004 and OHSUMED, while on TD2003 it performs poorly (for low $k$) or fairly well (for high $k$). 

We notice that there are four methods which only report results in two of the three datasets: the two SortNet versions are only reported on TD2003 and TD2004, while StructRank and C-CRF are only reported on TD2004 and OHSUMED. RankMatch compares similarly with SortNet and StructRank on TD2004, similarly to C-CRF and StructRank on OHSUMED and similarly to the two versions of SortNet on TD2003. This exhausts all the comparisons against the methods which have results reported in only two datasets. A fairer comparison could be made if these methods had their performance published for the respective missing dataset.

When compared to the methods which report results in all datasets, RankMatch entirely dominates their performance on TD2004 and is second only to IsoRank on OHSUMED (and performing similarly to QBRank).

These results should be interpreted cautiously; \cite{MinRob08} presents an interesting discussion about issues with these datasets. Also, benchmarking of ranking algorithms is still in its infancy and we don't yet have publicly available code for all of the competitive methods. We expect this situation to change in the near future so that we are able to compare them on a fair and transparent basis. 

\textbf{Consistency} In a second experiment we trained RankMatch with different training subset sizes, starting with $0.03 \cdot D(k) \cdot M$ and going up to $1.0 \cdot D(k) \cdot M$. Once again, we repeated the experiments with DORM using \emph{precisely} the same training subsets. The purpose here is to see whether we observe a practical advantage of our method with increasing sample size, since statistical consistency only provides an asymptotic indication. The results are plotted in Figure \ref{fig:err_shapes}-right, where we can see that, as more training data is available, RankMatch improves more saliently than DORM.

\textbf{Runtime} The runtime of our algorithm is competitive with that of max-margin for small graphs, such as those that arise from the ranking application. For larger graphs, the use of the sampling algorithm will result in much slower runtimes than those typically obtained in the max-margin framework. This is certainly the  benefit of the max-margin matching formulations of \cite{CaeMcACheLeSmo09,LeSmo07}: it is much faster for large graphs. Table 1 shows the runtimes for graphs of different sizes, both for exponential family and max-margin matching models. 
\begin{table}[t]
\label{tab:runtime}
\caption{Training times (per observation, in seconds) for the exponential model and max-margin. Runtimes for $M=3,4,5$ are from the ranking experiments, computed by full enumeration; $M=20$ corresponds to the image matching experiments, which use the sampler from \cite{HubLaw08}. A problem of size $20$ cannot be practically solved by full enumeration.}
\centering
\begin{tabular}{rrr}
M & exponential model & max margin\\
\hline
3 & 0.0006661 & 0.0008965\\
4 & 0.0011277 & 0.0016086\\
5 & 0.0030187 & 0.0015328\\
20 & 36.0300000 & 0.9334556
\end{tabular}
\end{table}

\subsection{Image Matching}
For our computer vision application we used a silhouette image from the Mythological Creatures 2D database\footnote{http://tosca.cs.technion.ac.il}.
We randomly selected 20 points on the silhouette as our interest points and applied shear to the image creating 200 different images. We then randomly selected $N$ pairs of images for training, $N$ for validation and 500 for testing, and trained our model to match the interest points in the pairs. In this setup,  
\begin{align}
x_{ij} = | \psi_i - \psi_j |^2, 
\end{align}
where $| \cdot |$ denotes the elementwise difference and $\psi_i$ is the Shape Context feature vector \cite{BelMal00} for point $i$.

\begin{figure*}[thb]
\vspace{-3mm}
\centerline{
\includegraphics[width=0.45\textwidth]{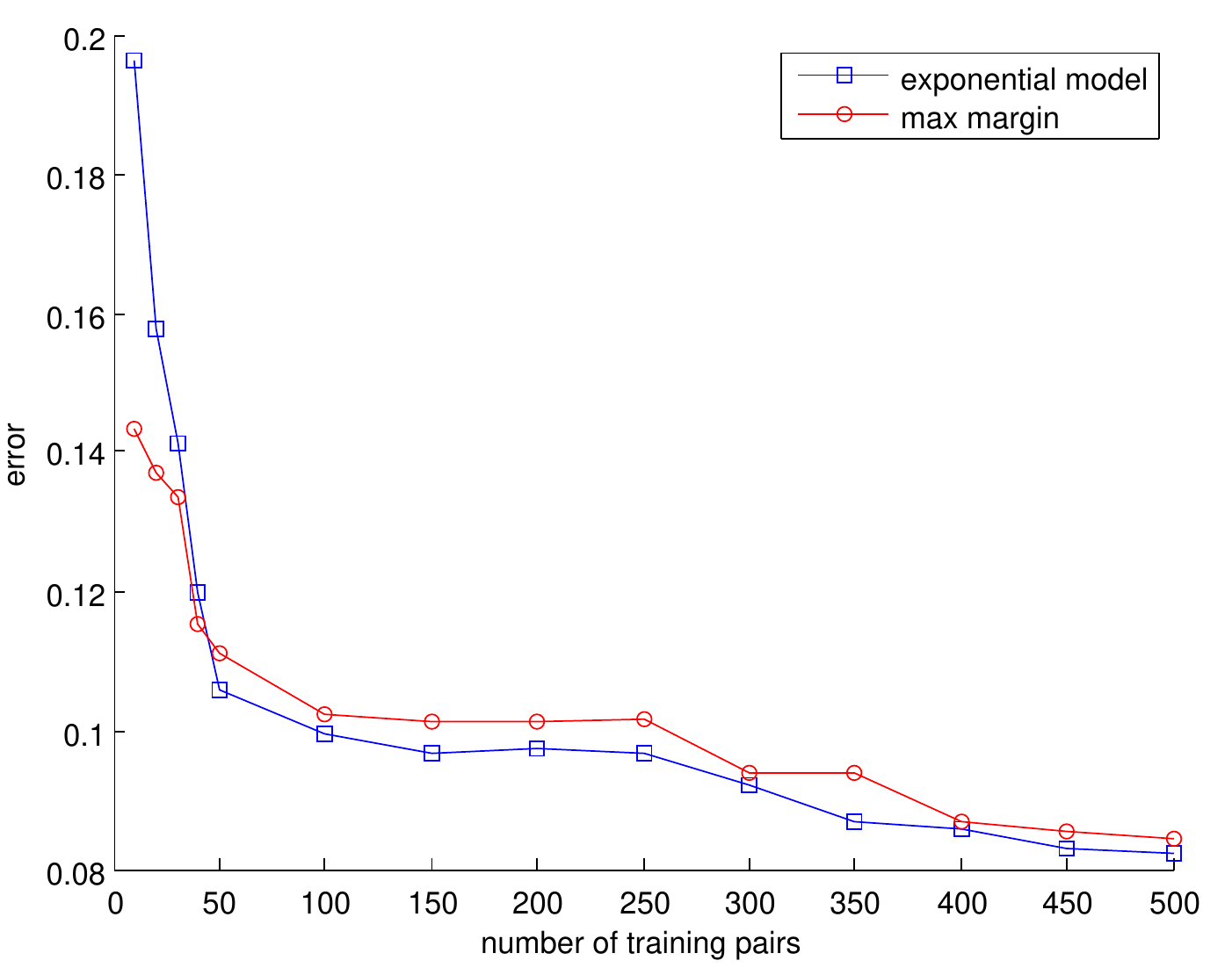}\hspace{1cm}
\includegraphics[width=0.45\textwidth]{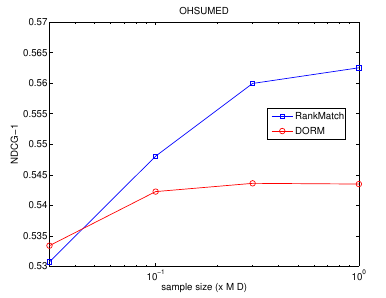} 
}
\caption{Performance with increasing sample size. Left: hamming loss for different numbers of training pairs in the image matching problem (test set size fixed to 500 pairs). Right: results of NDCG@1 on the ranking dataset OHSUMED. This evidence is in agreement with the fact that our estimator is consistent, while max-margin is not.}
\label{fig:err_shapes}
\end{figure*}

For a graph of this size computing the exact expectation is not feasible, so we used the sampling method described in Section \ref{sec:approx_exp}. 
Once again, the regularization constant $\lambda$ was chosen by cross-validation. Given the fact that the MAP estimator is consistent while the max-margin estimator is not, one is tempted to investigate the practical performance of both estimators as the sample size grows. However, since consistency is only an \emph{asymptotic} property, and also since the Hamming loss is \emph{not} the criterion optimized by either estimator, this does not imply a better large-sample performance of MAP in real experiments. In any case, we present results with varying training set sizes in Figure \ref{fig:err_shapes}-left. The max-margin method is that of \cite{CaeMcACheLeSmo09}. After a sufficiently large training set size, our model seems to enjoy a slight advantage. 

\section{Conclusion and Discussion}
We presented a method for learning max-weight bipartite matching predictors, and applied it extensively to well-known webpage ranking datasets, obtaining state-of-the-art results. We also illustrated--with an image matching application--that larger problems can also be solved, albeit slowly, with a recently developed sampler. The method has a number of convenient features. First, it consists of performing maximum-a-posteriori estimation in an exponential family model, which results in a simple unconstrained convex optimization problem solvable by standard algorithms such as BFGS.
Second, the estimator is not only statistically consistent but also in practice it seems to benefit more from increasing sample sizes than its max-margin alternative. Finally, being fully probabilistic, the model can be easily integrated as a module in a Bayesian framework, for example. The main direction for future research consists of finding more efficient ways to solve large problems. This will most likely arise from appropriate exploitation of data sparsity in the permutation group.

\section*{Appendix A}

For completeness we include a description of the sampling algorithm presented in \cite{HubLaw08}. The algorithm is an \emph{accept-reject} algorithm. The core idea of such an algorithm is very simple: assume we need to sample from a distribution $p$ in a given domain $\mathcal M$, but that such a task is intractable. Instead, we sample from a distribution $q$ in a \emph{superset} $\mathcal N$ of the original domain (in which sampling is easier), whose restriction to the original domain coincides with the original distribution: $q|_{_{\mathcal N}}=p$. We then only `accept' those samples that effectively fall within the original domain $\mathcal M$. Clearly, the efficiency of such a procedure will be dictated by (i) how efficient it is to sample from $q$ in $\mathcal N$ and (ii) how much mass of $q$ is in $\mathcal M$. Roughly speaking, the algorithm presented in  \cite{HubLaw08} manages to sample perfect matches of bipartite graphs such that both conditions (i) and (ii) are favorable. 

The reasoning goes as follows: the problem consists of generating variates $y\in \Ycal$ ($y$ is a match) with the property that $p(y)=w(y)/Z$, where $w(y)$ is the non-negative score of match $y$ and $Z=\sum_y w(y)$ is the partition function, which in our case is a permanent as discussed in Section 4.1. We first partition the space $\Ycal$ into $\Ycal_1, \dots, \Ycal_I$, where $\Ycal_i = \{y:y(1)=i\}$. Each part has its own partition function $Z_i = \sum_{y\in \Ycal_i} w(y)$. Next, a suitable upper bound $U(\Ycal_i)\ge Z_i$ on the partition function is constructed such that the following two properties hold:\footnote{See \cite{HubLaw08} for details.} 
\begin{align}
\nonumber
& \text{(P1)} \hspace{8mm} \sum_{i=1}^M  U(\Ycal_i)  \le U(\Ycal). \\
\nonumber
& \text{(P2)} \hspace{8mm} \text{If } |\Ycal_i|=1,  \text{ then } U(\Ycal_i)=Z_i=w(y).
\end{align}

That is, (i) the upper bound is super-additive in the elements of the partition and (ii) if $\Ycal_i$ has a single match, the upper bound \emph{equals} the partition function, which in this case is just the score of that match.

Now the algorithm: consider the random variable $\mathcal I$ where $p(\mathcal I=i) = U(\Ycal_i)/U(\Ycal)$. By (P1), $\sum_{i=1}^M p(i)\le 1$, so assume $p(\mathcal I = 0)=1-\sum_{i=1}^M p(i)$. Now, draw a variate from this distribution, and if $\mathcal I=i=0$, reject and restart, otherwise recursively sample in $\Ycal_i$.\footnote{Due to the self-reducibility of permutations, when we fix $y(1)=i$, what remains is also a set of permutations. We then sample $y(2),y(3)\dots y(M)$.} This algorithm either stops and restarts or it reaches $\Ycal_{\text{final}}$ which consists of a match, i.e.,~$|\Ycal_{\text{final}}|=1$. This match is then a legitimate sample from $p(y)$. The reason this is the case is because of (P2), as shown below. Assuming the algorithm finishes after $k$ samples, the probability of the match is the telescopic product
\begin{align}
\frac{U(\Ycal_{\mathcal I(1)})}{U(\Ycal)}\frac{U(\Ycal_{\mathcal I(2)})}{U(\Ycal_{\mathcal I(1)})}\dots \frac{U(\Ycal_{\mathcal I(k)})}{U(\Ycal_{\mathcal I(k-1)})}\stackrel{(P2)}{=}\frac{w(y)}{U(\Ycal)},
\end{align}
and since the probability of acceptance is $Z/U(\Ycal)$, we have 
\begin{align}
p(y) = \frac{w(y)/U(\Ycal)}{Z/U(\Ycal)} = \frac{w(y)}{Z},
\end{align}
which is indeed the distribution from which we want to sample. For pseudocode and a rigorous presentation of the algorithm, see \cite{HubLaw08}.

\end{document}